\documentclass[10pt,twocolumn,letterpaper]{article}

\usepackage[pagenumbers]{cvpr} %

\usepackage{graphicx, amsmath, amssymb, booktabs, enumitem, multirow}%

\usepackage[table]{xcolor}
\definecolor{citecolor}{HTML}{0071BC}
\definecolor{linkcolor}{HTML}{ED1C24}
\usepackage[pagebackref=false, breaklinks=true, colorlinks, citecolor=citecolor, linkcolor=linkcolor, bookmarks=false]{hyperref}

\usepackage[capitalize]{cleveref}
\crefname{section}{Sec.}{Secs.}
\Crefname{section}{Section}{Sections}
\Crefname{table}{Table}{Tables}
\crefname{table}{Tab.}{Tabs.}

\def\cvprPaperID{**}
\def\confName{****}
\def\confYear{****}

\newcommand{\z}{\mathbf{z}}
\newcommand{\R}{\mathbb{R}} 
\newcommand{\authorskip}{\hspace{2.5mm}}

\usepackage{color}
\definecolor{mygreen}{rgb}{0, 0.8, 0}

\begin{document}

\title{
\vspace{-1mm}\Large Modelling Latent Dynamics of StyleGAN using Neural ODEs\vspace{-3mm}}
\author{
Weihao Xia$^{1}$ \authorskip Yujiu Yang$^{2}$ \authorskip Jing-Hao Xue$^{1}$\\
$^{1}$Department of Statistical Science, University College London, UK\\ $^{2}$Tsinghua Shenzhen International Graduate School, Tsinghua University, China
}

\maketitle

\begin{abstract}
In this paper, we propose to model the video dynamics by learning the trajectory of independently inverted latent codes from GANs.
The entire sequence is seen as discrete-time observations of a continuous trajectory of the initial latent code, by considering each latent code as a moving particle and the latent space as a high-dimensional dynamic system.
The latent codes representing different frames are therefore reformulated as state transitions of the initial frame, which can be modeled by neural ordinary differential equations.
The learned continuous trajectory allows us to perform infinite frame interpolation and consistent video manipulation.
The latter task is reintroduced for video editing with the advantage of requiring the core operations to be applied to the first frame only while maintaining temporal consistency across all frames.
Extensive experiments demonstrate that our method achieves state-of-the-art performance but with much less computation.
Code is available at \small{\url{https://github.com/weihaox/dynode_released}}.

\end{abstract}

\section{Introduction}
\label{sec:intro}

GAN inversion~\cite{alaluf2021restyle,richardson2020encoding,xia2022gan} have recently been developed, allowing us to invert various kinds of images back into the latent space of pretrained GAN models.
We can then manipulate the attributes of images using latent editing methods~\cite{yao2021latent,shen2020interpreting,khrulkov2021latent}.
But these pretrained models, inversion methods and latent editing techniques are all being trained on and developed for images, limiting their applications in video processing.
Recent work~\cite{skorokhodov2022stylegan_v} suggests training a GAN for video synthesis.
Due to the lack of high-quality video datasets and the challenges of integrating an additional data dimension, video GANs have not been able to match their single-image counterparts.
Recent methods~\cite{yao2021latent,tzaban2022stitch} have shown that even equipped with an off-the-shelf and non-temporal StyleGAN, the elusive temporal coherency, an essential requirement for video tasks to meet,
can be achieved by maintaining that of the original video during inversion and latent editing processes.
These methods, however, require frame-by-frame operations.
The operations are almost the same across all frames, which makes us wonder: could these operations be applied just to the first frame, or specifically, 
could we change attributes for the entire video by only applying a latent editing method to the initial latent code?

Given a video, most existing GAN inversion methods invert consecutive frames separately without considering the temporal correlation between them. This leads to non-temporal latent codes, meaning that the independently-inverted latent codes become known but remain isolated in the latent space and their temporal relationships are not exploited. 
This paper approaches the question from a new perspective: we model the trajectory of (or equivalently, the temporal correlation among) the
latent codes inverted independently in a GAN's latent space.
More specifically, we propose a method called \texttt{DynODE}, to model the video \texttt{Dyn}amics by learning the trajectory of latent codes with neural ordinary differential equations (\texttt{ODE}s)~\cite{chen2018neural}.
Our work borrows the intuition from dynamical systems and treats the trajectory of a sequence as the solution to a first-order non-autonomous ODE.
By considering latent codes as moving particles and the latent space as a high-dimensional dynamical system, a video sequence is seen as the discrete-time observations of a continuous trajectory of the initial latent code.
The original latent space is therefore mapped into a dynamic space.
The latent codes corresponding to different frames are reformulated as state transitions of the first frame, which can then be modeled by using neural ODEs.

There are many advantages to the ODE-based approaches to our problem. 
First, they specify a continuous latent state. 
The neural ODEs are encouraged to learn the holistic geometry of the video dynamic space, allowing them to produce video frames at unseen timesteps.
It enables our method to perform \textit{continuous frame interpolation}.
This is especially helpful when the frames are irregular-sampled and when a temporally smooth video is required.
Furthermore, the learned trajectory facilitates \textit{consistent video manipulation} by operating primarily the first frame.
Our novel setting for consistent video manipulation only applies core operations to the first frame,
promoting the benefits of modeling such trajectories.
Some methods have been proposed for world- or time-consistent video manipulation in a frame-by-frame setting, but to our knowledge, this is the first time such a task has been accomplished by focusing primarily on the first frame, without repeating operations on all frames.
Our method can also be taken as a temporal counterpart of SinGAN~\cite{shaham2019singan}.
Trained using a single image, SinGAN produces assorted high-quality images of arbitrary sizes that semantically resemble the training image but contain new object configurations and structures.
Analogously, trained in a data-efficient way, our method learns the dynamics from a single video and produces arbitrary frames that contain temporally-consistent contents and motions. 
Extensive experiments on a wide range of datasets manifest that the proposed method learns the temporal correlation among the independently-inverted latent codes.
Experimental results also show that our approach is comparable to state-of-the-art face video editing methods that perform frame-by-frame editing, but requires much fewer operations.

The novelty and contributions of our work are summarized as follows.
1) We model the video dynamics by learning the temporal trajectory of the non-temporal latent codes with neural ODEs. 
2) We present a dynamical view of the latent space, in which video frames become discrete-time observations of a continuous trajectory of the initial latent code.
With this perspective, existing GAN inversion and latent editing techniques can be applied to video tasks in a different manner.
3) %
Our method allows for infinite frame interpolation and consistent video manipulation. 
The latter task is introduced under a novel setting where the core operations need only be applied to the first frame.
Compared with state-of-the-art methods that perform frame-by-frame editing, our approach achieves comparable performance with much less computation.

\begin{figure}[t!]
\centering
\includegraphics[width=\linewidth]{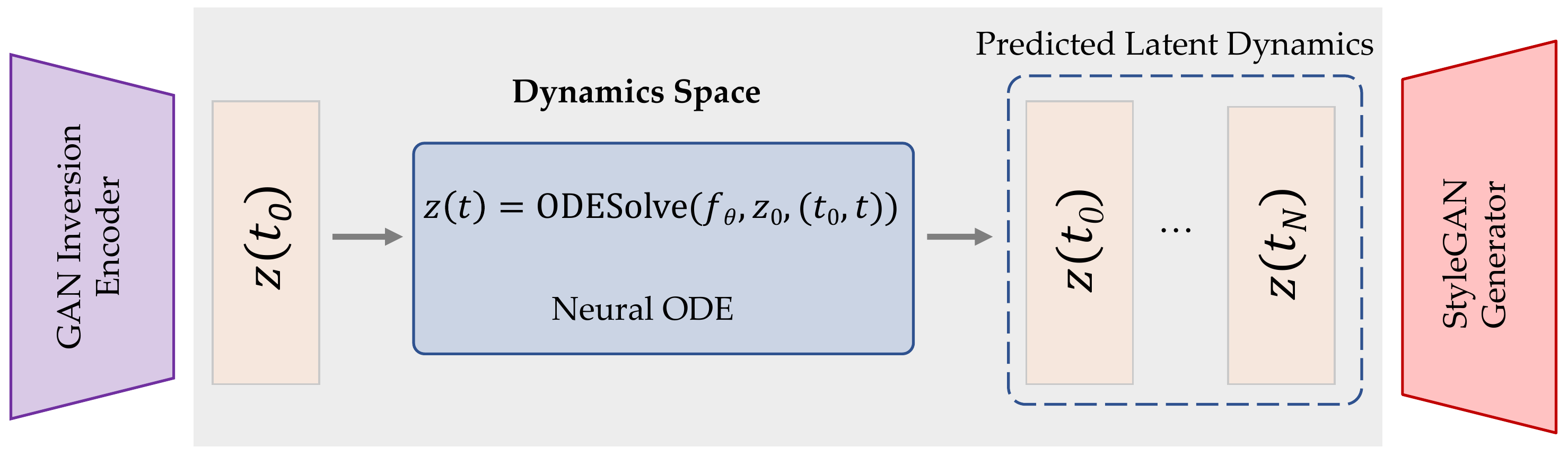}
\caption{Our goal with \texttt{DynODE} is to model the latent dynamics of StyleGAN using a neural ODE.
}
\label{fig:scheme}
\end{figure}

\section{Related Work}
\label{sec:related_work}

\noindent\textbf{Video Frame Interpolation.}
Video frame interpolation aims to synthesize intermediate frames between two consecutive video frames.
The key of video interpolation is to model the motion and generate the intermediate frame.
In~\cite{meyer2015phase}, Meyer~\etal~propose a phase-based video interpolation scheme that demonstrates impressive performance on videos with small displacements.
Niklaus~\etal~\cite{niklaus2017video} develop a kernel-based framework to predict a sampling kernel for every pixel and interpolate each pixel by convolution on corresponding patches of adjacent frames.
Many studies~\cite{liu2017video,jiang2018super} use optical flows to interpolate videos with large motions.
Different from existing methods, we present a data-efficient method that models video dynamics from a single video and allows for infinite frame interpolation.

\noindent\textbf{Consistent Video Manipulation.}
Despite steady progress in image editing, video editing remains difficult since such edits need to be applied consistently to all frames.
Kasten~\etal~\cite{kasten2021layered} present a way to edit natural videos by parameterizing them through layered neural 2D atlases.
Mallya~\etal~\cite{mallya2020world} convert semantic inputs into time- and view-consistent videos, based on projections from the point cloud to the camera.
More recently, some methods~\cite{yao2021latent,xu2021continuity,tzaban2022stitch} based on GAN inversion~\cite{alaluf2021restyle,richardson2020encoding,xia2022gan,bai2022high} and latent editing~\cite{yao2021latent,shen2020interpreting} have recently been proposed.
They follow a similar pipeline of altering the inverted latent code of each aligned-and-cropped frame.
The same operations are applied repeatedly to the latent code for each frame. We observe that such frame-by-frame processing contains lots of redundancy in operations.
By removing the redundant operations from the aforementioned methods, our framework achieves comparable performance at a considerably lower cost.
The produced video remains highly consistent in spite of applying the core operations to the first frame only.
In contrast with previous studies, we propose a method for editing video in a time-coherent manner without the need to apply redundant operations frame-by-frame.

\noindent\textbf{Video Applications of Neural ODEs.} 
Neural ODE~\cite{chen2018neural} and its variants have recently been explored for video generation.%
Kanaa~\etal~\cite{kanaa2019simple} combine a typical encoder-decoder architecture with Neural ODEs.
Park~\etal~\cite{park2020vid} propose an ODE convolutional GRU as the encoder for continuous-time video generation.
Unlike~\cite{kanaa2019simple,park2020vid}, we use a neural ODE network to model the temporal correlation among latent codes, which are derived separately by using existing GAN inversion algorithms.
Our work have significant differences from them in at least three key aspects: 
1) \textit{Task}. Their methods are proposed for video generation while our work is for consistent video interpolation and manipulation;
2) \textit{Mechanism}. They train an encoder-decoder with neural ODEs, while the only component of our method that needs training is the neural ODE network; 
and 3) \textit{Applicability}. Our method can handle videos at resolutions up to 1024$\times$1024. 
This presents a substantial improvement over previous methods, results of which are typically of much lower resolutions.
More importantly, our method is easily applicable to current video editing methods, freeing them from tedious frame-by-frame processing.

\begin{figure*}[thbp]
\centering
\includegraphics[width=\linewidth]{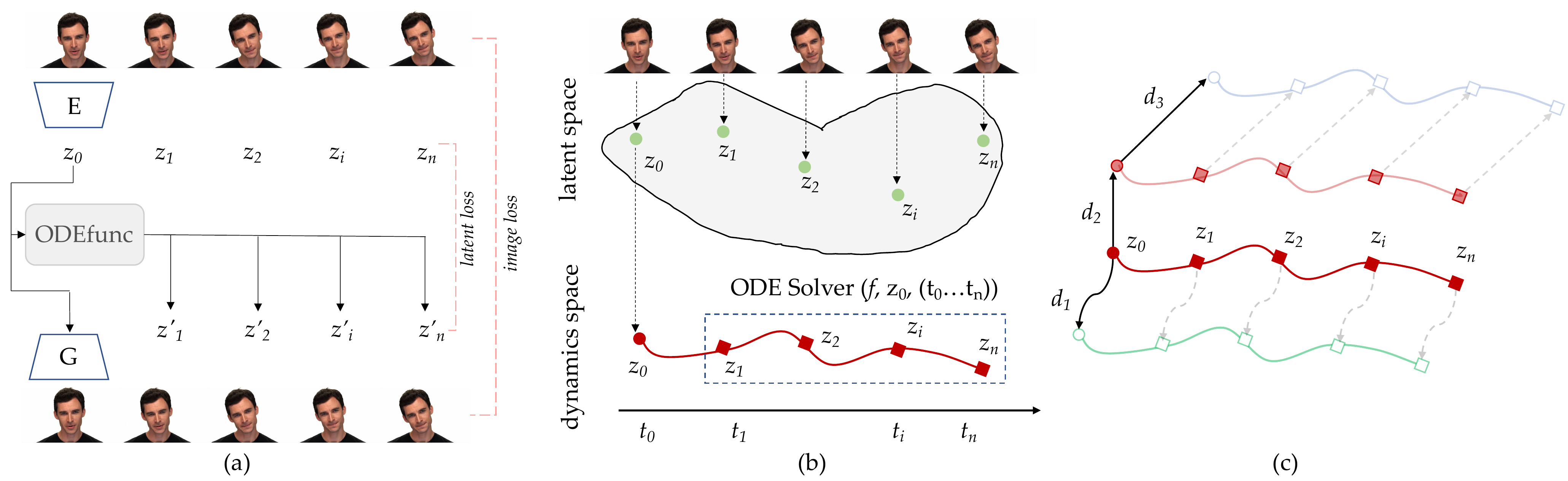}
\caption{{Framework of the proposed~\texttt{DynODE}.}
\textbf{(a)}~The neural ODE network ($\mathtt{ODEfunc}$, $f_{\theta}$) is trained to predict the subsequent latent states given the initial state by minimizing the losses in the latent, feature, and image spaces.
\textbf{(b)}~The original latent space is mapped into a dynamic space. 
These latent codes $\{\z_0, \cdots, \z_n\}$ in the latent space becomes discrete-time observations $\{{\z}(t_0), \cdots, {\z}(t_n)\}$ of a continuous trajectory of the initial latent state in the dynamic space.
The latent codes corresponding to different frames are reformulated as state transitions of the first frame, which can be modeled by using neural ODEs.
\textbf{(c)}~The desired video attributes could be edited by changing the first frame and extending such modifications to all subsequent frames.
These attributes could be altered by applying a direction, either nonlinear ($d_1$) or linear ($d_2$ and $d_3$), to the initial latent code.
}
\label{fig:framework}
\end{figure*}

\section{Method}
\label{sec:method}

\subsection{Preliminary}
\label{sec:method-preliminary}

\noindent\textbf{GAN Inversion} aims to invert a given image back into the latent space of a pretrained GAN model so that the image can be faithfully reconstructed from the inverted code by the generator.
The generator of an unconditional GAN learns the mapping $\mathcal{G}: \mathcal{Z} \to \mathcal{X}$. 
When $\z_1, \z_2 \in \mathcal{Z}$ are close in the $\mathcal{Z}$ space, the corresponding images $x_1, x_2 \in \mathcal{X}$ are visually similar. 
GAN inversion maps data $x$ back to latent representation $\z^*$ or, equivalently, finds an image ${x^*}$ that can be entirely synthesized by the well-trained generator $\mathcal{G}$ and remain close to the real image $x$.
Formally, denoting the signal to be inverted as $x \in \R^{n}$, the well-trained generative model as $\mathcal{G}: \R^{n_{0}} \to \R^{n}$, and the latent vector as $\z \in \R^{n_{0}}$, GAN inversion studies the following problem:
\begin{equation}
\z^*=\underset{\z}{\arg \min } \ \ell(\mathcal{G}\left(\z), x\right),
\label{eqn:def}
\end{equation}
where $\ell(\cdot)$ is a distance metric in the image or feature space, and $\mathcal{G}$ is assumed to be a feed-forward neural network. 
With the inverted $\z^*$, we can obtain the original and manipulated images. 
These pretrained GAN models, the inversion methods, and the latent editing techniques are trained on and developed for images, limiting their applications in video processing.
\textit{Instead of developing alternative GAN architectures or inversion methods for videos, we model the trajectory of the isolated latent codes to apply existing inversion techniques to video tasks.}

\noindent\textbf{Neural ODEs} are a family of continuous-time models which define a hidden state $h(t)$ to be the solution to an ODE initial-value problem:
\begin{align}
    \dot{h}(t)= \frac{d h(t)}{dt} = f_\theta\left(h(t), t;\theta\right) \quad
    \textit{s.t.}
    \quad h(t_0) = h_0.
\end{align}
The function $f_\theta$ specifies the dynamics of the hidden state, using a neural network with parameters $\theta$. 
$t \in[0, T]$ is time and $h(t) \in \mathbb{R}^{d}$. 
The hidden state $h(t)$ is defined at all times, and can be evaluated at any desired time step by using a numerical ODE solver denoted as $\mathtt{ODESolve}$:
\begin{align}\label{eq:latent_ode}
h_0,\dots,h_N = \mathtt{ODESolve}\left(f_\theta, h_0, (t_0,\dots,t_N)\right),
\end{align}
where $(t_0, \dots, t_N)$ are time points where $h(t)$ is evaluated. The gaps between consecutive time points $t_i$ are not necessarily equal.
Neural ODEs specify $h(t)$ as a continuous function over time, even though it is evaluated at discrete time points $(t_0,\dots,t_N)$. 

\subsection{DynODE}
\label{sec:dynode}

Given the inverted codes ${\z_0, \cdots, \z_n}$ corresponding to each frame in the sequence ${c_0, \cdots, c_n}$,  our objective is to model the trajectory or dynamics of a video sequence in the latent space. This specifically targets the latent dynamics of StyleGAN, illustrated in \cref{fig:objective}.
If we treat the latent space as a dynamic system, change in latent codes along a certain direction is analogous to the motion of particles.
Our work borrows intuition from dynamical systems and treats the trajectory of the entire sequence as the solution to a first-order non-autonomous ODE.
By considering the latent space as a dynamical system and modeling latent codes as moving particles, the entire sequence can be viewed as discrete-time observations of a continuous trajectory of the initial latent code $\z_0$.
The subsequent latent codes become state transitions of the initial one.
Therefore, the dynamics of the entire video is actually determined by the first frame and its trajectory.

We consider the dynamics of a state $\z(t)$ in the phase space $\Omega$ ($=\mathbb{R}^{2n}$) of a dynamical system.
These non-temporal latent codes $\{\z_0, \cdots, \z_n\}$ become observations $\{{\z}(t_0), \cdots, {\z}(t_n)\}$ of a motion trajectory of $\z_0$ at specified times $t_0, \cdots,t_n$.
$\z(t_0)$ represents the initial state equal to $\z_0$.
This trajectory can be treated as the solution to a non-autonomous dynamical system determined by
\begin{equation} 
\label{eq:1}
\frac{d\z}{dt} = f\left(\z(t), t\right)
\quad \text{for } t \in \mathbb{R},\, \z \in \Omega,
\end{equation}
where $f: \Omega \times \mathbb{R}  \mapsto T\Omega$ is assumed to be continuous, and $T\Omega$ is the tangent space.
By approximating the differential with an estimator $f_{\theta}\simeq f$, where $f_{\theta}$ is a $\theta$-parameterized neural network,
the neural ODEs allow to model the evolution across time of such a dynamical system and learn the dynamics or trajectories from relevant data.
For an arbitrary time $t_i$, the $\mathtt{ODESolve}$ computes a numerical approximation of the integral of the dynamics from the initial time value $t_0$ to $t_i$ by~\cref{eq:1} that is equal to
\begin{equation}
\label{eq:3}
\begin{aligned}
\z_{i}= \tilde{\z}(t_i) 
& = \mathtt{ODESolve} \left(f_{\theta}, \z_{0},(t_{0}, t_{i})\right) \\
& \simeq \z_{0}+\int_{t_{0}}^{t_{i}} f_{\theta}\left(\z(t), t\right) dt = \z(t_{i}), \\
\text{where} \quad \z_0  & =\mathcal{E}(c_0), \quad \hat{c}_{i} =\mathcal{D}(\z_i).
\end{aligned}
\end{equation}
$\tilde{\z}(t_i)$ is a prediction of ${\z}(t_i)$ using a neural ODE network.

\cref{fig:framework} (a) shows the procedure of our framework\footnote{Despite a similar illustration with~\cite{kanaa2019simple}, the methodological and training mechanisms are fundamentally different.}.
For a source video consisting of $N$ frames $\left\{x_{i}\right\}_{i=1}^{N}$, we first obtain the preprocessed frames $\left\{c_{i}\right\}_{i=1}^{N}$ after a cropping-and-alignment step.
We then use an off-the-shelf GAN inversion method, which is denoted by $\mathcal{E}$, to produce their latent inversion $\left\{\z_{i}\right\}_{i=1}^{N}=\left\{\mathcal{E}\left(c_{i}\right)\right\}_{i=1}^{N}$ for all frames. 
The reconstructed image can be generated from latent code $\tilde{\z}_{i}$ with a generator $\mathcal{G}$ by $\mathcal{G}(\tilde{\z}_{i})$.
These latent codes are then used to train a neural ODE network ($\mathtt{ODEfunc}$, $f_{\theta}$), which aims to predict the subsequent frames.
Specifically, we randomly sample a small batch with the same size, $\z_0, \cdots, \z_n$ ($n<N$), for each iteration.
Given the initial state $\z_0$, the neural ODE network is trained to predict the known observations, $\z_1, \cdots, \z_n$, at the corresponding times by minimizing the losses in the latent, feature, and image spaces.
The obtained latent codes are then fed into the respective generators in order to produce images.
Notice that the generators for the latent codes of the entire video might be different, for instance, if the generators are fine-tuned for each latent code as in~\cite{roich2021pivotal}.
The same applies to the GAN inversion module $\mathcal{E}$. 
The accurate description for $\mathcal{E}$ and $\mathcal{G}$ in~\cref{fig:framework}~(a) should be $\mathcal{E}_1, \cdots, \mathcal{E}_N$ and $\mathcal{G}_1, \cdots, \mathcal{G}_N$, respectively.
One for each is illustrated in the figure for simplicity.
This means that our approach is not reliant on any specific methods of GAN inversion, latent editing, or any other related techniques, and thus can be implemented as a plug-and-play module for video dynamic modeling.

\cref{fig:framework} (b) depicts the mapping from the spatial latent space to the temporal dynamic space. 
By considering the latent space as a high-dimensional dynamical system and latent codes as moving particles, 
the non-temporal latent codes
in the latent space becomes discrete-time observations
of a continuous trajectory of the initial latent state in the dynamic space.
The latent codes corresponding to different frames are reformulated as state transitions of the first frame.
We model the dynamics of these isolated latent codes in the latent space and learn a temporal trajectory in the dynamic space using a neural ODE function.
Once trained, dividing the time interval by $k$, the neural ODE network produces the states at the specified time points in the form of $\z_0,\dots,\z_k = \mathtt{Solve}(f_\theta, \z_0, (t_0,\dots,t_k))$.
Furthermore, with this learned trajectory, as illustrated in \cref{fig:framework}~(c), it is possible to perform consistent video manipulation, meaning to change the desired attributes of the target subject in a video by merely altering the first frame and consistently extending those modifications to all subsequent frames.
As aforementioned, the editing in the initial frame could be easily done by applying a desired direction to its latent code.
The direction (marked as the solid arrow lines in the figure), either nonlinear ($d_1$) or linear ($d_2$ and $d_3$), could be obtained by the off-the-shelf latent editing approaches~\cite{shen2020interpreting,yao2021latent} or some recent language-driven image manipulation methods~\cite{patashnik2021styleclip,xia2021tedigan,xia2021towards}. 
This is thought to be equivalent to applying particular directions from latent editing methods (dashed arrow lines) to all latent codes observed at certain time points.

\begin{figure*}[!t]
\centering
\includegraphics[width=0.92\linewidth]{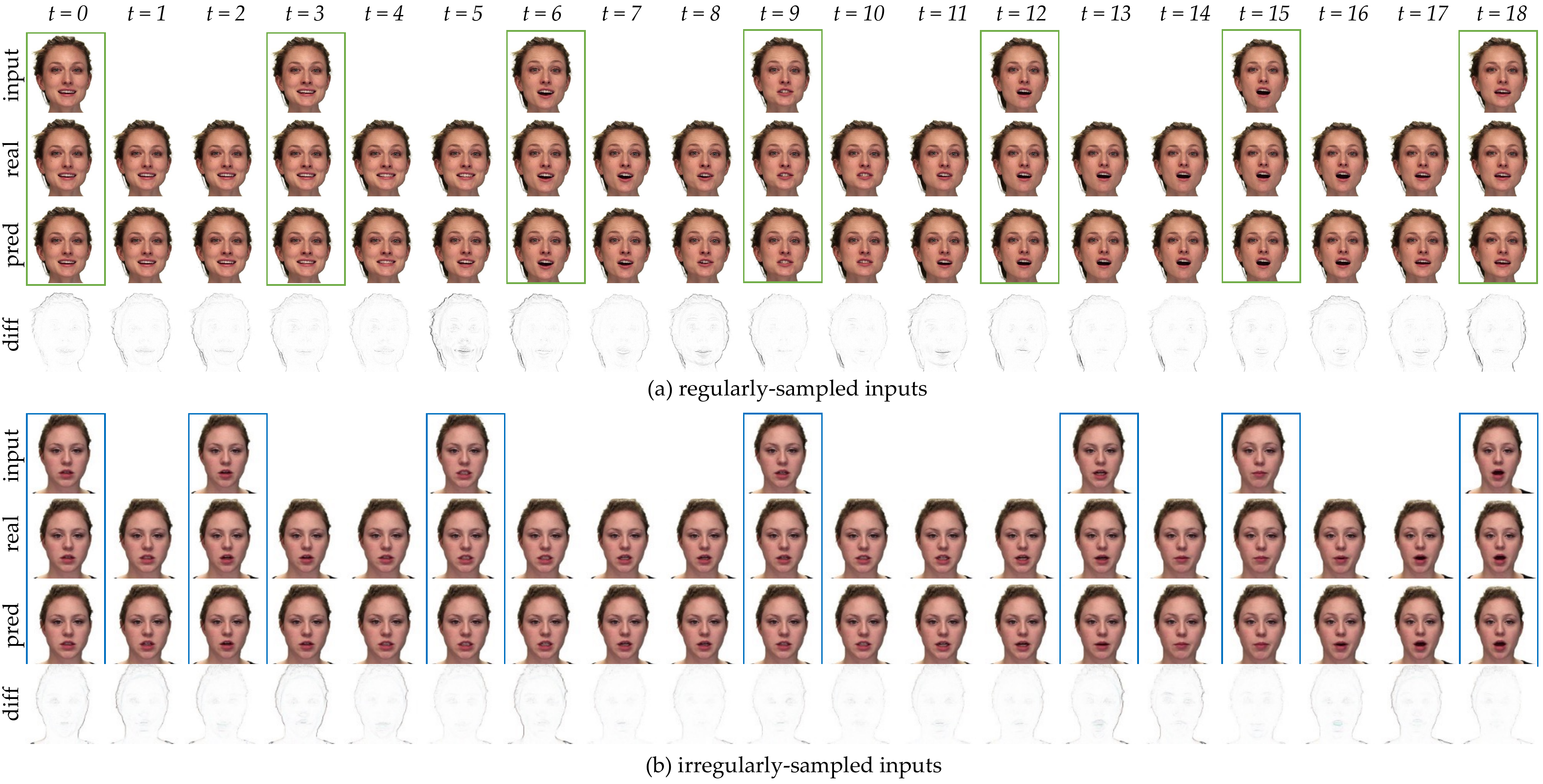}
\caption{Qualitative results of dynamic modeling at both observed and unobserved times.
We sample frames at regular and irregular time intervals and compare the predicted frames with the actual ones at both observed and unobserved time points.
}
\label{fig:ode_results}
\end{figure*}

\begin{table*}[!t]
\centering
\rowcolors{2}{gray!0}{gray!5}
\scalebox{0.92}
{
\begin{tabular}{lcccccccc}
\toprule
Dataset 
& \multicolumn{2}{c}{\texttt{Face}} & \multicolumn{2}{c}{\texttt{Scene}} & \multicolumn{2}{c}{\texttt{Bird}}
& \multicolumn{2}{c}{\texttt{Isaac3D}}\\\cmidrule(lr){2-3}\cmidrule(lr){4-5}\cmidrule(lr){6-7}\cmidrule(lr){8-9}
Metric &MSE $\downarrow$ &SSIM $\uparrow$ &MSE $\downarrow$ &SSIM $\uparrow$ &MSE $\downarrow$ & SSIM $\uparrow$ &MSE $\downarrow$ &SSIM $\uparrow$ \\
\midrule
Observed &6.752 &98.9 &22.956 &97.6 &25.407 &98.1 &15.655 &99.2 \\
Unobserved &35.397 &93.2 &48.323 &92.4 &55.368 &90.3 &27.651 &96.5 \\
\bottomrule
\end{tabular}
}
\caption{Quantitative evaluation of dynamic modeling on four datasets. The performance of dynamic modeling here is defined as the reconstruction quality of predicted images at both observed and unobserved time points. Metrics are reported in MSE $\downarrow$ ($\times$e-3) and SSIM $\uparrow$. $\downarrow$ means lower is better, while $\uparrow$ means the opposite.}
\label{tab:quan_dynamic}
\end{table*}

\section{Experiments}
\label{sec:experiment}

This section describes datasets, models, and results in terms of three aspects: video dynamic modeling, infinite frame interpolation, and consistent video manipulation.

\begin{figure*}[thbp]
\centering
\includegraphics[width=0.85\linewidth]{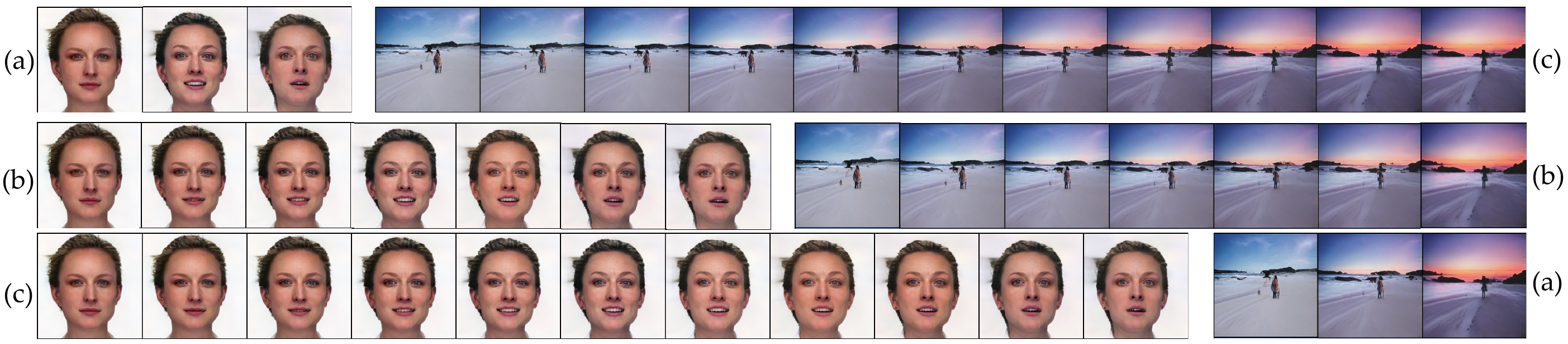}
\caption{Results of continuous frame interpolation for talking faces and outdoor natural scenes. 
Based on given frames in (a), our method can generate in-between video frames in diverse time intervals.}
\label{fig:interpolation}
\end{figure*}

\begin{figure}[thbp]
\centering
\includegraphics[width=0.82\linewidth]{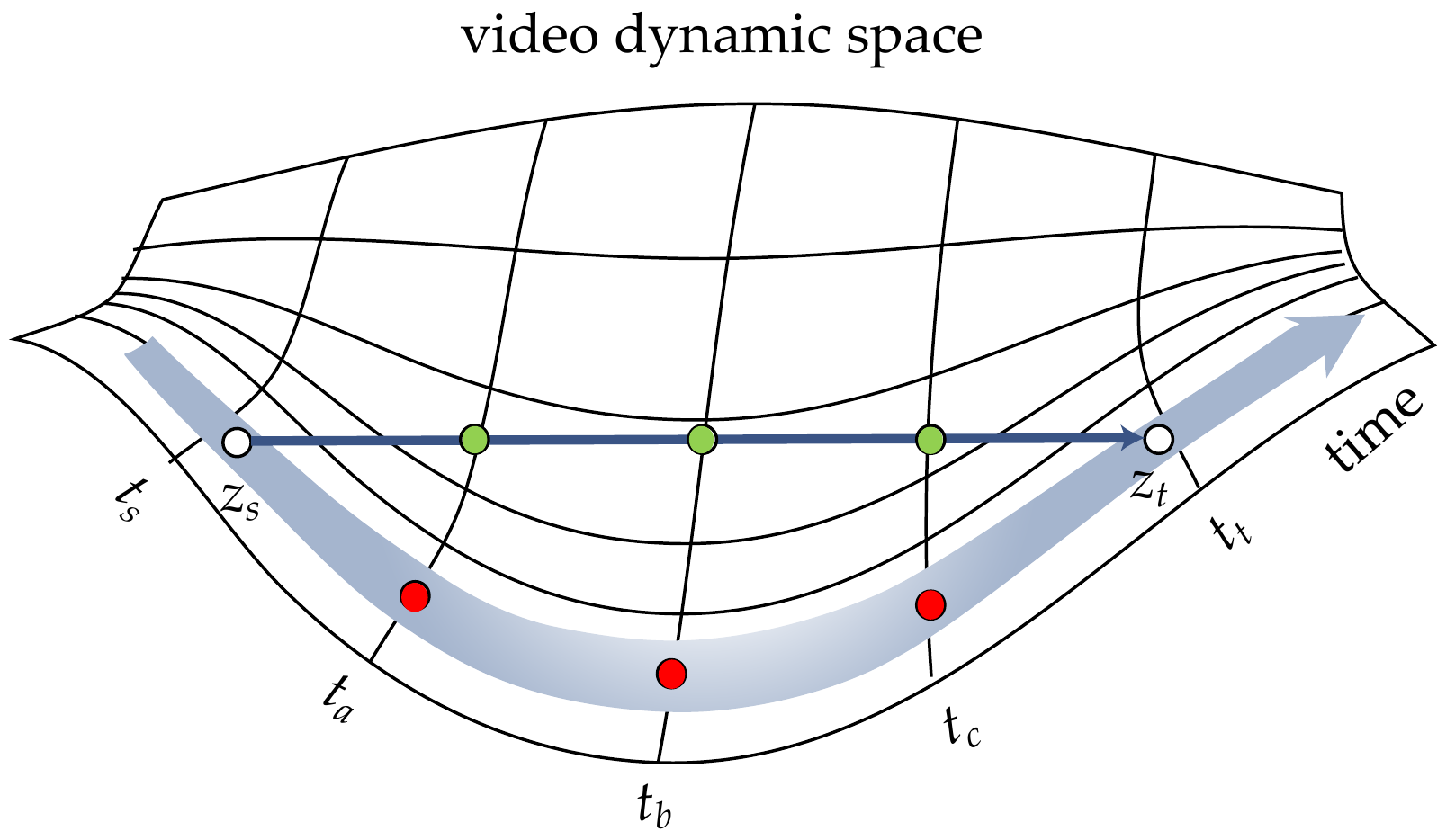}
\caption{
In the video dynamic space, the direct morphing creates the intermediate states (\textcolor{mygreen}{green}) by accumulating the differences between the two states at $t_s$ and $t_t$.
In contrast, neural ODEs estimate in-between states (\textcolor{red}{red}) at unseen times by accounting for the holistic geometry of the video dynamics. These states produce  time-oriented and motion-coherent frames.
}
\label{fig:dynamics_space}
\end{figure}

\subsection{Experimental Settings}

\noindent\textbf{Pretrained models and datasets.}
Given their widespread popularity, most of the GAN inversion and latent editing methods focus on StyleGANs~\cite{karras2019style,karras2020analyzing}.
Therefore, we use StyleGAN2 ~\cite{karras2020analyzing} as the pretrained model in our experiments.
Our method is not dependent on StyleGANs and can be applied to any GAN model. 
The latent codes are obtained by inverting each frame into $\mathcal{W^+}$ space of StyleGAN2~\cite{karras2020analyzing}. Other latent spaces~\cite{xia2022gan}, \eg~$\mathcal{Z}$ or $\mathcal{W}$ space, are also supported.
Experiments are conducted on several categories of publicly available datasets to demonstrate the effectiveness of our proposed method.

\begin{itemize}[leftmargin=*, topsep=1pt, itemsep=1pt]

\item \textbf{\texttt{Face}.} 
The StyleGAN2 model is trained on LLHQ~\cite{karras2019style} at a resolution of 1024$\times$1024. The real face videos are collected from publicly available talking-head datasets~\cite{wang2021one,livingstone2018ryerson} or downloaded from \textsc{YouTube}.

\item \textbf{\texttt{Scene}.} The StyleGAN2 is trained using Place365~\cite{zhou2017scene} at the resolution of 256$\times$256. 
We use~\cite{khrulkov2021latent} to generate temporally-consistent frames by editing the time-varying attributes,~\eg,~\textsc{night},~\textsc{dawndusk}, and~\textsc{sunrisesunset}.
We also download real videos of outdoor natural scenes from \textsc{YouTube} and obtain latent codes of sampled frames by direct optimization. 

\item \textbf{\texttt{Bird}.}
The StyleGAN2 model is trained using CUB-200-2011 dataset~\cite{wah2011caltech} at the resolution of 256$\times$256. We collect bird videos from \textsc{YouTube}.

\item \textbf{\texttt{Isaac3D}.} 
The StyleGAN2 model is trained using Isaac3D~\cite{nie2020semi} with a resolution of 128$\times$128. 
The dataset contains nine factors of variation, such as background color, object shape, robot movement, and camera height. 
As it is a synthetic dataset, we generate consecutive frames by moving the robot or camera.
\end{itemize}

\noindent\textbf{Implementation details.}
We implement the proposed method in PyTorch on an Nvidia GeForce RTX 2080.
The neural ODE function is parameterized by a Multi-Layer Perceptron (MLP).
The parameters in the neural ODE function are optimized using the Adam optimizer~\cite{kingma2014adam}. 
We train the neural ODE network with 5000 gradient descent steps with a learning rate of $0.01, \beta_{1}=0.9, \beta_{2}=0.999$, and $\epsilon=1 e^{-8}$.
We use a adaptive-step \textsc{dopri5} (Runge-Kutta~\cite{dormand1980family} of order 5 of Dormand-Prince-Shampine) as the default ODE Solvers.

\noindent\textbf{Performance metrics.}
The performance metrics for quantitative comparisons are divided into two groups.
To evaluate dynamic modeling, we use pixel-wise Mean Square Error (MSE) and Structural Similarity Index (SSIM)~\cite{TIP2004ImageWang}.
For comparisons of consistent video manipulation, we focus on evaluating the temporal coherence of edited videos, which is measured by the identity similarity between the frame pairs as in~\cite{tzaban2022stitch}.
To evaluate temporal consistency of facial identity, we use two metrics, temporally-local (TL-ID) and temporally-global (TG-ID) identity preservation introduced in~\cite{tzaban2022stitch}, and a variant of the Average Content Distance (ACD)~\cite{tulyakov2018mocogan}.
We use Fréchet Video Distance (FVD)~\cite{unterthiner2018towards} to evaluate the quality of motion. 

\subsection{Dynamic Modeling} 
\label{sec:exp_dynamic}
We conduct experiments to evaluate the performance of video dynamic modeling.
Neural ODEs are expected to learn the temporal properties of a trajectory, allowing it to approximate the actual states at the observed time points and even those between observations.
Thus, the performance of dynamic modeling can be understood as the reconstruction quality of predicted images.
Specifically, we sample frames at regular and irregular time intervals and compare the predicted frames with the actual ones at both observed and unobserved time points.
\cref{fig:ode_results} shows the sampled frames of the original videos and the predicted results.
\cref{tab:quan_dynamic} shows the performance of dynamic modeling characterized by the reconstruction quality.
Considering that images would be perfectly fitted if given enough time, we set a fixed number of iterations to maintain a reasonable running time.
The quantitative evaluation on four categories demonstrates that our method not only models the video dynamics for the known observations but also generalizes to the unobserved time. 
This capability indicates that neural ODEs are able to learn the entire geometry of video dynamic rather than simply remembering given observations, enabling them to accept irregularly sampled frames or create video frames at unknown times.

There are some ways to analyze the spatio-temporal dynamics of videos in addition to ODE-based models.
Recent studies~\cite{park2020vid,kidger2020neuralcde} have demonstrated that they are not as effective as ODE-based models.
The RNN-based models, for example, assuming fixed time intervals, are limited to learn the representations only at the observed times.
These methods can barely handle datasets collected from wild environments with irregular time intervals and missing states.

\begin{figure*}[!th]
\centering
\includegraphics[width=0.82\linewidth]{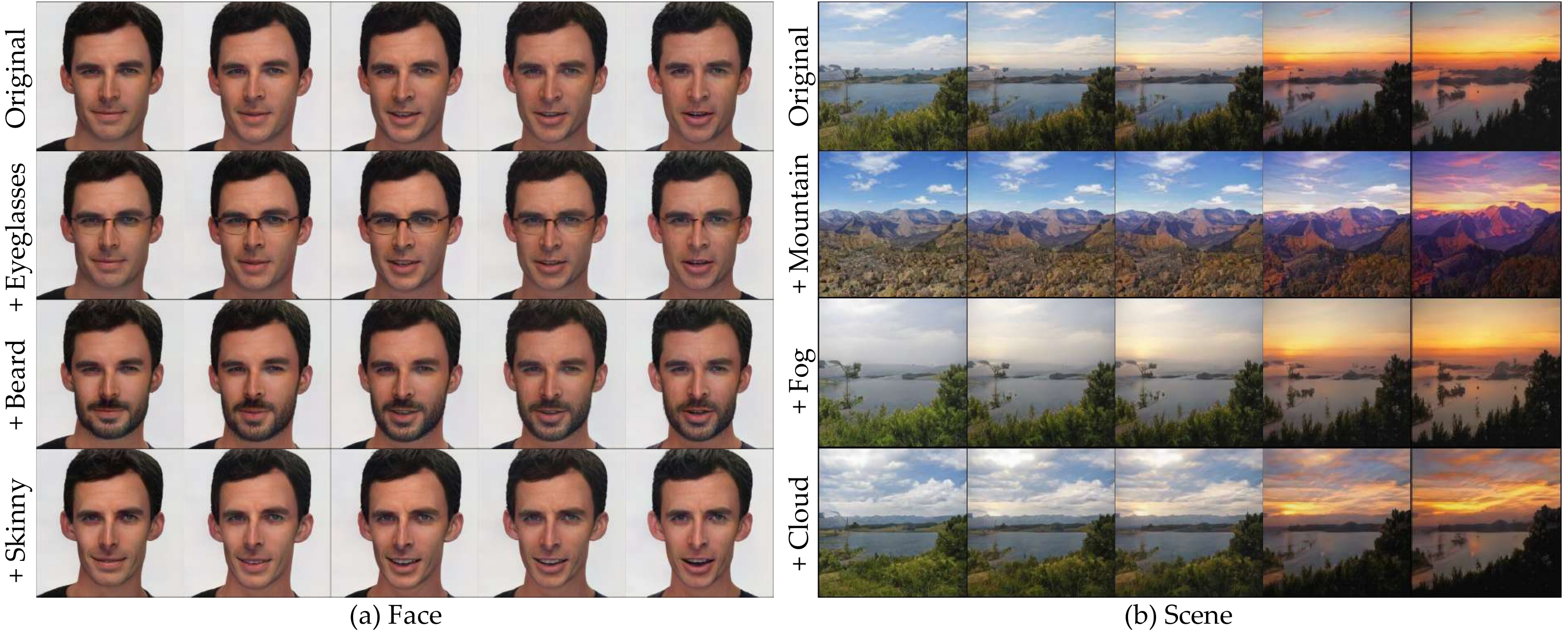}
\caption{
Results of consistent video manipulation for talking heads and outdoor natural scenes. 
Our method changes the desired attributes of the entire video by altering the initial frame and extending such modifications to the entire sequence, without the need to apply redundant operations to every frame.
The manipulated frames of the entire video show identical video dynamics and maintain temporal coherence, even when the facial identity in the first frame appears to have drifted after editing.
}
\label{fig:consistent}
\end{figure*}

\subsection{Continuous Frame Interpolation} 
\label{sec:exp_interpolation}
The learned neural ODE specifies $z(t)$ as a continuous function over time, which facilitates infinite frame interpolation. 
It enables our framework to interpolate non-existent frames within the time interval from $t = 0$ to $t = N$ at arbitrary time steps.
Once trained, dividing the time interval $[0, N]$ by $k$, the neural ODE network produces the in-between states at the corresponding time points in the form of $\z_0,\dots,\z_k = \mathtt{Solve}(f_\theta, \z_0, (t_0,\dots,t_k))$.
The intrinsic properties of neural ODEs allow to achieve such infinite frame interpolation at a constant memory cost even when the frames are irregularly sampled or partially observed, 
which is the case that their RNN-based counterparts are often struggling to deal with~\cite{kidger2020neuralcde,rubanova2019latent}.
This is particularly helpful when a temporally smooth video is required.
\cref{fig:interpolation} shows the results of frame interpolation for talking heads and outdoor natural scenes at arbitrary time steps. 
While the generators are trained on image datasets and not specifically tuned for the video, the predicted frames still exhibit high consistency across time.

It should be noted that intermediate video frames can also be generated by simply blending two adjacent latent codes.
This procedure, often called image morphing in literature, is to fuse two images by interpolating their latent codes, which also presents a continuous process.
Given $z_s$ and $z_t$, a series of semantically meaningful images can be generated following $z^*=z_s+\alpha(z_t-z_s)$ , where $\alpha$ is a scale between $0$ and $1$.
The term $(z_t-z_s)$ can also be considered as a direction, similar to those discovered by latent editing methods~\cite{yao2021latent,khrulkov2021latent,shen2020interpreting}.
As opposed to the trajectory learned by neural ODEs, which could be viewed as the \textit{temporal} directions, these are \textit{spatial} or \textit{manipulatable} directions for altering the semantics.
\cref{fig:dynamics_space} (adopted from~\cite{park2020vid}) illustrates a video dynamic space from $t_s$ to $t_t$.
The 2D instead of 1D structure of such space indicates the stochasticity of the trajectory between the two observed time points.
The difference between direct morphing and ours is obvious in the video dynamic space.
The direct morphing creates the intermediate states (\textcolor{mygreen}{green}) at $t_a$, $t_b$, and $t_c$ by accumulating the differences between $\z_s$ and $\z_t$, without considering the geometry structure of the video dynamic space.
The obtained states may fall outside of the video dynamic space and thus produce frames that contain spatial changes instead of temporal motions.
In contrast, our method estimates in-between states (\textcolor{red}{red}) at unknown times by accounting for the holistic geometry of the video dynamics. 
These states produce time-oriented and motion-coherent frames.
The trajectory obtained from our method, illustrated as the \textcolor{blue}{blue} arrow curve, fits closely with the video dynamic space and indicates the intrinsic motion of the video.

Due to the inherent limitations of current GANs~\cite{karras2020analyzing}, our method is suited more to videos with a specific category and may not perform well on existing video frame interpolation benchmarks that include different objects or scenes.
The existing frame interpolation methods~\cite{xu2019quadratic,jiang2018super,liu2017video,niklaus2017video} could not be trained for specific categories due to the lack of such high-quality video datasets.
As a result, we do not conduct comparisons of video interpolation. The comparison will be applicable if either constraint is lifted in the future. 

\subsection{Consistent Video Manipulation}
\label{sec:exp_editing}

\noindent\textbf{Baselines.}
We compare our method with several state-of-the-art GAN inversion based video editing baselines~\cite{tzaban2022stitch,yao2021latent}.
These recently-proposed methods follow a pipeline that consists of three key steps: pre-processing (inverting each cropped and aligned frame into the latent space); attribute manipulation (editing images by
employing off-the-shelf latent based semantic editing techniques), and seamless cloning (blending the modified faces with the original input frames using~\cite{perez2003poisson}). 
To maintain the inherent temporal consistency of the original video, Tzaban~\etal~\cite{tzaban2022stitch} further include an Pivotal Tuning Inversion (PTI) module~\cite{roich2021pivotal} to produce faithful inversions by fine-tuning the generator and introduce a stitching-tuning operation that further tunes the generator to provide spatially-consistent transitions.
Xu~\etal~\cite{xu2021continuity} propose to improve inversion quality by utilizing a sequence of frames instead of a single image. 
We compare with~\cite{xu2021continuity} to see how this video inversion scheme contributes to video editing.

\begin{figure*}[t!]
\centering
\includegraphics[width=0.82\linewidth]{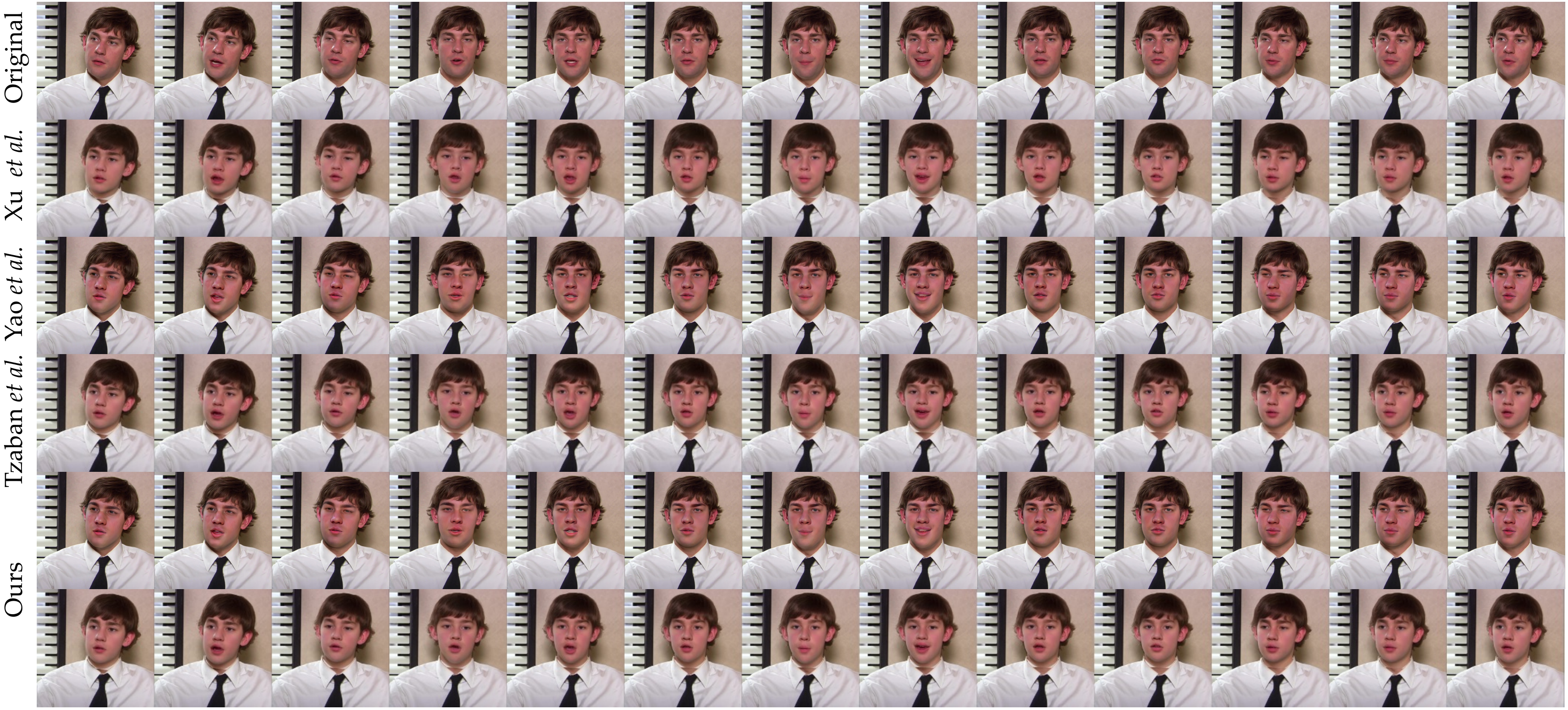}
\caption{Comparison with recent face video editing methods, including Yao~\etal~\cite{yao2021latent}, Tzaban~\etal~\cite{tzaban2022stitch}, and Xu~\etal~\cite{xu2021continuity}.
{The last two rows are obtained through plug-and-play integration of our method into existing video editing pipelines~\cite{yao2021latent,tzaban2022stitch} respectively.} 
}
\label{fig:qual_vid_editing}
\end{figure*}

\begin{table*}[!t]
\centering
\rowcolors{2}{gray!0}{gray!5}
\scalebox{0.92}
{
\begin{tabular}{lccccc}
\toprule
&Yao~\etal~\cite{yao2021latent} &Xu~\etal~\cite{xu2021continuity} &Tzaban~\etal~\cite{tzaban2022stitch} &Ours (\cite{yao2021latent}/\cite{tzaban2022stitch}) \\ \midrule
TL-ID $\uparrow$ &0.957 &0.969 &\underline{0.989} &0.965/\textbf{0.991} \\
TG-ID $\uparrow$ &0.839 &0.851 &\textbf{0.912} &0.843/\underline{0.907} \\
ACD $\downarrow$ &1.352 &1.485 &\textbf{0.846} &1.331/\underline{0.929} \\
FVD $\downarrow$ &582.4 &632.7 &\textbf{352.9} &522.3/\underline{378.2} \\
\bottomrule
\end{tabular}
}
\caption{Quantitative comparison with recent face video editing methods.
We reported results of identity preservation metrics and FVD. $\uparrow$ means higher is better, while $\downarrow$ means the opposite.}
\label{tab:quan_vid_editing}
\end{table*}

\noindent\textbf{Qualitative evaluation.}
We demonstrate the effectiveness of our method on a range of in-the-wild videos gathered from popular publicly available content.
These include diverse and challenging scenarios characterized by complex backgrounds and considerable movement.
\cref{fig:consistent} shows results of consistent video editing on talking heads and outdoor natural scenes.
The target attribute is edited by employing off-the-shelf latent-based semantic editing techniques (\ie,~\cite{yao2021latent} for faces and~\cite{khrulkov2021latent} for others) on the first frame.
Our method changes the desired attributes of the entire video by altering the initial frame and extending such modifications to the entire sequence, without the need to apply redundant operations to every frame.
The edited video displays identical dynamics and maintains temporal coherence, even when the first frame after editing shows drifted facial identities in some cases (\eg~\textsc{Skinny}).
The drifted identities result from the employed latent-based editing process~\cite{yao2021latent} and can be resolved by using alternative methods.

The comparison results of facial attribute editing on videos are shown in~\cref{fig:qual_vid_editing}.
We compare several state-of-the-art GAN inversion based video editing baselines~\cite{xu2021continuity,yao2021latent,tzaban2022stitch}.
These results demonstrate highly temporal coherence across all the frames. 
The reason for this is that the temporal consistency of the source video is maintained during inversion and latent editing.
The reconstructed video would be predisposed towards temporal consistency if the image-based GAN inversion methods can faithfully reconstruct each frame, even without considering the temporal relationship between adjacent frames.
Furthermore, despite minor inversion inconsistencies introduced by the inversion module, especially those of the backgrounds, the final results would still be smooth in these regions.
The blending process only stitches the masked facial regions back to the original images and the inconsistent backgrounds are not involved.
The video-based inversion method
~\cite{xu2021continuity}, on the other hand, does not provide extra feasibility for the task of video editing compared with its image-based counterparts.
The last two rows are results from implementing our method as a plug-and-play module for dynamic modeling in two existing face video editing pipelines~\cite{yao2021latent} and~\cite{tzaban2022stitch}, respectively.
Note that our results are obtained by applying core operations only to the first frame, freeing the aforementioned video editing methods from repeating tedious and redundant operations on all frames.
The desired video attributes could be edited by altering the first frame and extending such changes to the entire sequence, as we expected, without deteriorating the temporal consistency across all the frames. 

\noindent\textbf{Quantitative evaluation.}
Since the results of all compared methods are generated by the same generator, the evaluation on image quality using the popular metric,~\eg, FID or LPIPS, are not very meaningful.
Instead, we evaluate the temporal coherence and the motion quality of edited videos. 
The temporal coherence is measured by the identity similarity between the frame pairs.
TL-ID computes the identity similarity between adjacent video frames. 
TG-ID refers to the identity similarity between all possible pairs of video frames. 
We also evaluate the temporal consistency of facial identity using a variant of ACD~\cite{tulyakov2018mocogan}.
We use FVD~\cite{unterthiner2018towards} to evaluate the quality of motion.

The quantitative evaluation results are shown in~\cref{tab:quan_vid_editing}.
Our method achieves state-of-the-art performance when embedded as a plug-and-play dynamic modeling module in two face video editing frameworks~\cite{yao2021latent} and~\cite{tzaban2022stitch}, respectively.
In contrast to other video editing studies, our method avoids repeating operations on every frame, without sacrificing editing quality or temporal consistency.

\section{Discussion} 
\label{sec:conclusion}

\noindent\textbf{Limitations and future directions.} Our framework has several limitations. For example, we use the simplest implement of the neural differential equations to show their potential applications. 
From the vast variants, we can provide the dynamic adjustment to the trajectory based on future observations using the neural controlled differential equations (CDEs)~\cite{kidger2020neuralcde}, or introduce the stochasticity using the neural stochastic differential equations (SDEs)~\cite{li2020scalable}. %
Furthermore, since the dynamics is modeled from one single video, the learned trajectory is deterministic, which contradicts the stochastic nature of the time-varying videos.
The problem can be addressed by training an encoder on a large-scale video dataset. The learned encoder would be capable of stochastic video prediction based on the first frame.

\noindent\textbf{Conclusion.} In this paper, we present \texttt{DynODE}, a method to model video dynamics by learning the trajectory of independently-inverted latent codes using neural ODEs.
Our method estimates time-oriented and motion-coherent frames at unseen timesteps by accounting for the holistic geometry of the video dynamic space.
Such a design enables continuous frame interpolation and consistent video manipulation, freeing video editing from tedious and redundant frame-by-frame processing.
Extensive experiments on a wide range of datasets show that our method improves upon prior state-of-the-art methods.

{\small
\bibliographystyle{ieee_fullname}
\bibliography{reference}
}

\end{document}